\documentclass[10pt,twocolumn,letterpaper]{article}

\usepackage{cvpr}              % To produce the CAMERA-READY version
\usepackage{graphicx}
\usepackage{amsmath}
\usepackage{amssymb}
\usepackage{booktabs}
\usepackage{multirow}
\usepackage{array}

\usepackage{algorithm}
\usepackage{algorithmic}

\usepackage{hyperref}
\hypersetup{hidelinks,
	        colorlinks=true,
	        pdfstartview=Fit,
	        breaklinks=true}

\usepackage[capitalize]{cleveref}
\crefname{section}{Sec.}{Secs.}
\Crefname{section}{Section}{Sections}
\Crefname{table}{Table}{Tables}
\crefname{table}{Tab.}{Tabs.}

\def\cvprPaperID{*****} % *** Enter the CVPR Paper ID here
\def\confName{CVPR}
\def\confYear{2022}

\begin{document}

\title{Improve Ranking Correlation of Super-net through Training Scheme from One-shot NAS to Few-shot NAS}

\author{Jiawei Liu\footnotemark[1],  Kaiyu Zhang\footnotemark[1],  Weitai Hu\footnotemark[1], and Qing Yang \\
Du Xiaoman Financial, Beijing, China\\
{\tt\small \{liujiawei, zhangkaiyu, huweitai,  yangqing\}@duxiaoman.com}
}

\maketitle
\renewcommand{\thefootnote}{\fnsymbol{footnote}} 
\footnotetext[1]{Equal Contribution. Listing order is random.}

%%%%%%%%% ABSTRACT
\begin{abstract}
The algorithms of one-shot neural architecture search (NAS) have been widely used to reduce computation consumption. However, because of the interference among the subnets in which weights are shared, the subnets inherited from these super-net trained by those algorithms have poor consistency in precision ranking. To address this problem, we propose a step-by-step training super-net scheme from one-shot NAS to few-shot NAS. In the training scheme, we firstly train super-net in a one-shot way, and then we disentangle the weights of super-net by splitting them into multi-subnets and training them gradually. Finally, our method ranks 4th place in the CVPR2022 3rd Lightweight NAS Challenge Track1. Our code is available at \href{https://github.com/liujiawei2333/CVPR2022-NAS-competition-Track-1-4th-solution}{https://github.com/liujiawei2333/CVPR2022-NAS-competition-Track-1-4th-solution}.
\end{abstract}

%%%%%%%%% BODY TEXT
\section{Introduction}
\label{sec:intro}
\thispagestyle{plain}
Neural architecture search (NAS) aims to automatically design neural network architectures. One-shot NAS is a kind of widely-used NAS method which utilizes a super-net subsuming all candidate architectures (subnets) to implement NAS function. All subnets directly inherit their weights from the super-net which is only trained once. The advantage of one-shot NAS  is it can provide a large number of subnets for a single training session at a low training cost.  

Methods like SPOS\cite{spos} and FairNAS\cite{fairnas} independently connect all candidates of the operation to construct a super-net, sampling one or more subnets and training them in each training iteration. On this basis, OFA\cite{once-for-all} constructs the super-net by entangling the weights of candidates with inclusion relations together, and the super-net is trained by the progressive shrinking method and knowledge distillation. Based on the super-net constructed by OFA, BigNAS\cite{bignas} trains the super-net at one time through sandwich rule and several tricks and no longer needs to fine-tune the subnet. However, the mentioned methods suffer from a common disadvantage which will cause interference among the subnets, so that the accuracy ranking of subnets is inconsistence with when they are trained from scratch independently.  

Some methods are attempting to improve the ranking correlation. Pi-NAS\cite{pinas} reduces the consistency problem to feature offset and parameter offset, then solves the problem through a self-supervised learning scheme of cross-path training super-net. Landmark Regularization\cite{landmark} takes the accuracy of a few standalone trained subnets as regularization to guide the training of super-nets. The 3rd place solution of CVPR2021 Lightweight NAS Challenge track1\cite{guanone} shows the influence on different aspects, including super-net construction, super-net training, and sampling on ranking correlation, which draws empirical conclusions through a large number of experiments. The 1st place solution of CVPR2021 Lightweight NAS Challenge track1\cite{improving} maps the super-net to the high-capacity super-net in stages and adds narrow branches of the super-net. Few-shot NAS\cite{zhao2021few} randomly selects different candidates as one group to split the search space, thus generating multiple sub-super-nets. GM-NAS\cite{GM-NAS} uses the gradient matching method to propose a few-shot NAS judgment benchmark for space segmentation. K-Shot NAS\cite{k-shot} trains k super-nets and samples subnets by combining the weights of these super-nets.  

Beyond that,  two studies bring some new thinking. \cite{Deeper-insights} designs a small search space and find a phenomenon: the ranking correlation varies greatly in different epochs and is even unstable in different iterations of the same epoch. \cite{How-does} also finds several interesting conclusions: the increase of epoch could hardly improve the ranking correlation, and the accuracy of top-K subnets has little relationship with ranking correlation. In addition, the ranking correlation is strongly correlated with the search space.

In this paper, we propose a step-by-step training scheme for super-net from one-shot NAS to few-shot NAS. We believe that reducing the degree of weight entanglement among subnets is the key to improving the consistency of super-net, so we gradually split the weight of a one-shot super-net so that make it decoupled to a few-shot super-net, and then we follow knowledge distillation and sandwich rule to gradually train the super-net. From a one-shot super-net to a few-shot super-net, the weight of the super-net is inherited step by step. In addition, we use an adaptive learning rate to alleviate the problem of different frequencies of module sampling in the few-shot super-net.

\section{Proposed method}
\label{sec:intro}

\subsection{Search Space}
Different from CVPR2021 Lightweight NAS Challenge Track1, which only searches the number of channels, this track in 2022 searches for the number of layers of the network and the number of channels at each layer. It would be very difficult to construct a super-net like what SPOS\cite{spos} and FairNAS\cite{fairnas} have done, so we construct that according to the ways of OFA\cite{once-for-all} and BigNAS\cite{bignas} with weight entanglement. The super-net is constructed based on resnet48, which contains 4 stages. The number of searchable channels is the product of the basic channel and scaling ratio. The scaling ratio in all stages is [1.0, 0.95, 0.9, 0.85, 0.8, 0.75, 0.7]. The number of searchable blocks is from 2 to 8. The details of search space in every stage are shown in Table \ref{tab:a}.

\begin{table}[!htbp]
  \centering
  \begin{tabular}{c@{}c@{}c@{}}
    \toprule
    Stage & Block & Basic channels \\
    \midrule
    1 & [2,3,4,5] & 64 \\
    2 & [2,3,4,5] & 128  \\
    3 & [2,3,4,5,6,7,8] & 256 \\
    4 & [2,3,4,5] & 512 \\ 
    \bottomrule
  \end{tabular}
  \caption{The search space of CVPR2022 Lightweight NAS challenge track1}
  \label{tab:a}
\end{table}

\subsection{Training super-net}
For the one-shot super-net, we follow the pipeline of BigNAS\cite{bignas}. In detail, we first construct the super-net based on the search space of $S$ with weights $\boldsymbol{\theta}$, and then we train the largest sub-network $S_{max}$ with the cross-entropy loss $\mathcal{L}_{CE}$. Finally, we distill the smallest sub-network $S_{min}$ and two sub-networks sampled randomly respectively with the Kullback–Leibler (KL) loss (distance) $\mathcal{L}_{KL}$ between the logit of them and the one of $S_{max}$. Algorithm \ref{alg:alg} shows the training process.  

As for the few-shot super-net, to alleviate the interference among the subnets whose weights are shared, the proposed method follows the way of few-shot NAS, which disentangles the shared weights by splitting the super-net into multi-subnet during the training process. In this section, we introduce the split way for the resnet-like search space.  

\begin{figure*}[htb]
\centering
\includegraphics[scale=0.4]{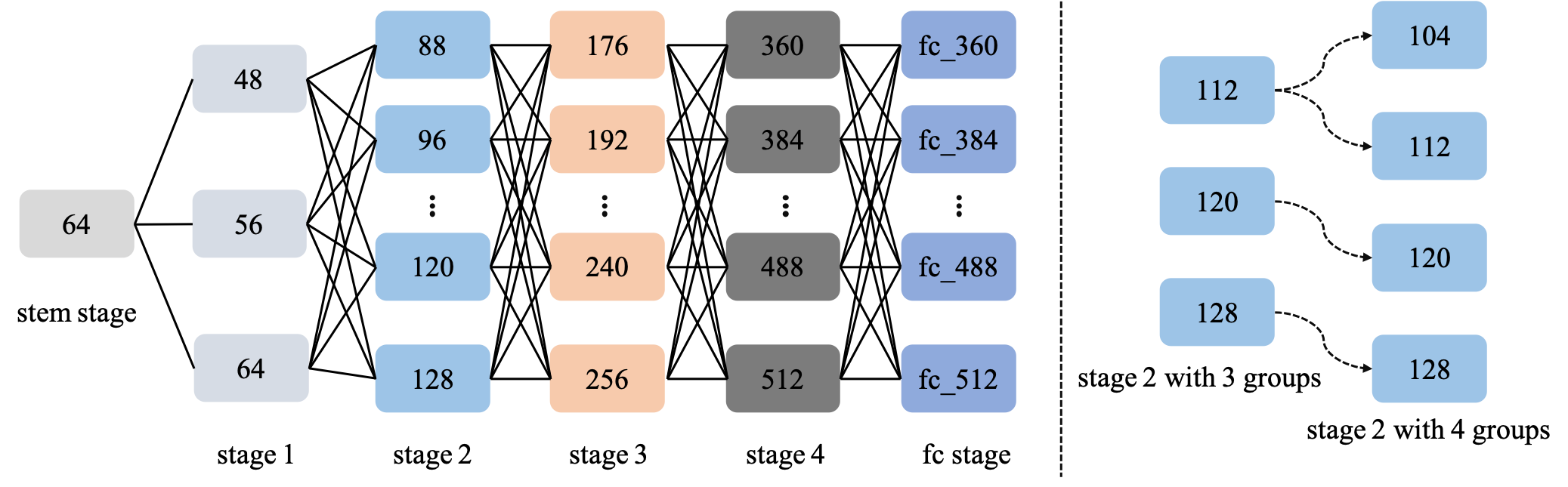}
\caption{\textbf{Left:} The super-net architecture with the largest group numbers. The box with number represents one group with the number as the output-feature channel. The solid-black line means the flow of features. From stem stage to full-connection stage, each stage includes 1, 3, 6, 7, 7, 7 groups respectively. \textbf{Right:} The inheriting method. Taking stage 2 as an example, the dashed-black arrow represents the source of the weights for the 4 groups in the second stage. Especially, the weights  of the second convolution layer in the group with 104 output channels is a replica from the weights of the same layer containing the former 104 channels in the previous group with 112 output channels.
}
\label{fig:1}
\end{figure*}

There are two kinds of channel number of convolution layers that need to be searched in one block, in which the second convolution layer controls the channel number of the output features. Because of the existence of identity mapping in the block of the resnet, the channel number of the output features is the same in every block in the same stage. In other words, the second channel number in different blocks in the same stage is fixed in one search process. Based on this analysis, we divide the search space following the dimensionality of the number in the second convolution layer. Specifically, the candidates for the second channel number are 88, 96, 104, 112, 120, and 128 for the second stage of this search space, whose basic channel number is 128. Instead of retaining only one group based on the vanilla one-shot NAS, we extend the number of the group up to the number of the candidates, that is to say, we divide the second stage into six groups, each of which contains the same architecture except for the channel number of the second convolution layer. We apply this split way to all stages except for the stem stage. The left of Figure \ref{fig:1} shows the details of the architectures of all stages.  

According to the aforementioned ways, we first train the super-net with one group by Algorithm \ref{alg:alg} until convergence (the same as the one-shot way), and then the super-net with two groups is trained following the same algorithm with the weights inheriting from the ones of the super-net with one group (The details of weights inheriting can be seen in the right of Figure \ref{fig:1}). We repeat the training process until the number of the group increases to maximum.  

\begin{algorithm}[!htbp]
    \renewcommand{\algorithmicrequire}{\textbf{Input:}}
	\renewcommand{\algorithmicensure}{\textbf{Output:}}
	\caption{One-shot Super-net Training} 
	\label{alg:alg} 
	\begin{algorithmic}[1]
		\REQUIRE super-net $S$, weights $\boldsymbol{\theta}$, total training step $T$, loss $\mathcal{L}_{CE}$ and $\mathcal{L}_{KL}$. 
		\ENSURE the optimal weights $\boldsymbol{{\theta}^*}$. 
		\STATE fix random seed.
		\STATE random initialize $\boldsymbol{\theta}$.
		
		\FOR{$t \leftarrow 1 , T$}
            \STATE set all gradients to zeros for super-net $S$.
            \STATE train the largest sub-net $S_{max}$ with loss $\mathcal{L}_{CE}$.
            \STATE calculate gradients for super-net $S$ based on $\mathcal{L}_{CE}$.
                \FOR{$k \leftarrow 1 , 3$}
                    \IF{$k \neq 3$}
                        \STATE random sample sub-net $S_{sample}$;
                    \ELSE
                        \STATE sample the smallest sub-net as $S_{sample}$;
                    \ENDIF
                    \STATE distill the $S_{sample}$ with the outputs of $S_{max}$;
                    \STATE calculate gradients based on $\mathcal{L}_{KL}$.
                \ENDFOR
                \STATE update $\boldsymbol{\theta}$ by accumulated gradients.
            \ENDFOR
            \STATE return the optimal weights $\boldsymbol{{\theta}^*}$.
	\end{algorithmic} 
\end{algorithm}

\section{Experiments}
\label{sec:intro}
\subsection{Results of one/few-shot super-net}
We firstly train the one-shot super-net for 100 epochs with batch size 128 and apply a stochastic gradient descent optimizer with the momentum of 0.9. The learning rate (LR) is decreased from an initial value of 0.1 to 0 with a linear-warmup cosine learning rate decay strategy, and the warmup epoch is 5. For the largest subnet, the weights are regularized with a weight decay of 0.00001, and the dropout probability is 0.2. The data augmentation includes brightness and contrast changes with a probability of 0.5, random rotation within 15 degrees, and random clipping of 224*224.  

The training configurations of few-shot NAS are mostly the same as that of one-shot NAS, except that epoch is set to 20 and there is no warm-up for the learning rate. In addition, different network modules are assigned different learning rates according to their probability of being sampled, to ensure the same order of magnitude in the training times. As shown in Table \ref{tab:b}, taking group 3 as an example, the LR of the module with channel number 112 in stage 1 is 4 times of the basic LR. LR of the module with channel number 232 in stage 2 and channel number 464 in stage 3 is 5 times of basic LR, and the learning rate of the rest modules is the basic learning rate.

\begin{table}[!htbp]
  \centering
  \renewcommand\tabcolsep{5.0pt}
  \begin{tabular}{cccc}
    \toprule
    Group & Stage & Channels & LR \\
    \midrule
    \multirow{3}{*}{2} & 1   & 64  & 2x \\
      & 2   & 128     & 5x \\
      & 3,4 & 256,512 & 6x \\
    \hline
    \multirow{2}{*}{3} & 2   & 112 & 4x \\
      & 3,4 & 232,464 & 5x \\
    \hline
    \multirow{2}{*}{4} & 2   & 104     & 3x \\
      & 3,4 & 216,432 & 4x \\
    \hline
    \multirow{2}{*}{5} & 2   & 96      & 3x \\
      & 3,4 & 208,408 & 4x \\
    \hline
    \multirow{2}{*}{6} & 2   & 88      & 3x \\
      & 3,4 & 192,384 & 4x \\
    \bottomrule
  \end{tabular}
  \caption{In different groups of few-shot NAS, the LR corresponding to different modules}
  \label{tab:b}
\end{table}

We selected the corresponding model at each training super-net stage to calculate the score, as shown in Table \ref{tab:c}.

\begin{table}[!htbp]
  \centering
  \renewcommand\tabcolsep{5.0pt} 
  \begin{tabular}{cccc}
    \toprule
    Stage & Group & Epoch & Pearson Coeff. \\
    \midrule
    One-shot &   & 80 & 0.80821 \\
    \hline
    \multirow{6}{*}{Few-shot} & 2  & 11  & 0.81008 \\
             & 3 & 17 & 0.81759 \\
             & 4 & 19 & 0.81547 \\
             & 5 & 18 & 0.81463 \\
             & 6 & 4 & \textbf{0.81830} \\
             & 7 & 11 & 0.81521 \\
    \bottomrule
  \end{tabular}
  \caption{The ranking correlation of subnets with weights inherited from super-net}
  \label{tab:c}
\end{table}

\subsection{Ranking changes while training}
\begin{figure}[!htbp]
\centering
\includegraphics[scale=0.2]{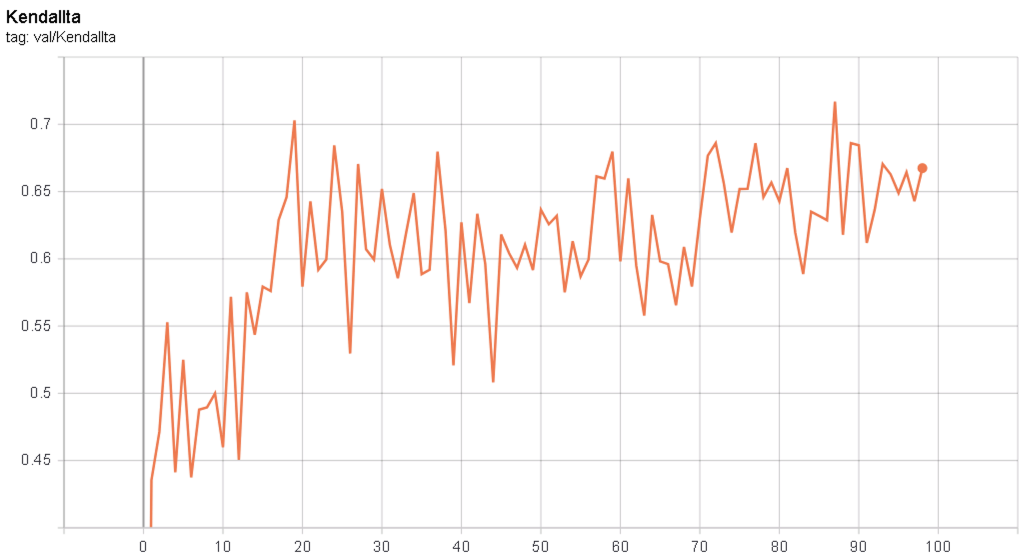}
\caption{Kendall tau changes over epochs.}
\label{fig:2}
\end{figure}

As \cite{Deeper-insights} said, the ranking does not increase steadily during training as accuracy does. Therefore, it is important to select the most appropriate epoch for the super-net. However, the large test set and a large number of subnets to be evaluated make it time-consuming to calculate rankings on each epoch. To solve this problem, we standalone trained 30 subnets with the part of valid datasets (the amount is $\frac{1}{20}$ of the original datasets') to obtain the target accuracy, and then used the same datasets to evaluate the accuracy of the 30 subnets inherited from the weight of the current super-net after each epoch when training the super-net. Therefore, we obtained the change curve of Kendall tau with the epoch as shown in Figure \ref{fig:2}, which shows that the ranking is rising but very unstable. The super-net with late epoch and large Kendall tau is selected as the measured model.

\subsection{Ablation on Architecture Sampling Strategies}
We discuss the different ranking performances of super-nets based on different architecture sampling strategies in this paragraph. To reduce the test time, we evaluate the super-net trained with the one-shot method on the part of datasets mentioned before. Table \ref{tab:d} shows the results of the uniform sample and fair sample, in which the former is better than the latter obviously, opposite to the results of \cite{fairnas}. This may be caused by the wrong usage of the fair samples. In \cite{fairnas}, it requires the same amount of candidates in the search space, which is not satisfied strictly in this search space.

\begin{table}[!htbp]
  \centering
  \renewcommand\tabcolsep{5.0pt} 
  \begin{tabular}{cc}
    \toprule
    Sample  & Pearson Coeff. \\
    \midrule
    Uniform & 0.6093 \\
    Fair    & 0.42489 \\
    \bottomrule
  \end{tabular}
  \caption{The comparison of different sampling strategies}
  \label{tab:d}
\end{table}

\section{Conclution}
In this paper, with the training scheme in combination with one-shot NAS and few-shot NAS, we improve the ranking consistency of the subnets whose weights are inherited from the ones of super-net. In detail, we first train the super-net with the one-shot method, and then to disentangle the weights, we extend the stages of super-net from one group up to the maximum along with the weights inhering. Finally, the different stages in super-net with different groups use the adaptive LR for sufficient training. In addition, we investigate the relationship between the consistency of subnets and the epochs, as well as the influences of the sample strategies of subnets on the ranking correlation.

%%%%%%%%% REFERENCES
{\small
\bibliographystyle{ieee_fullname}
\bibliography{egbib}
}

\end{document}